\documentclass[10pt]{article} 
\usepackage[accepted]{tmlr}

\usepackage{amsmath,amsfonts,bm}

\def\eqref#1{equation~\ref{#1}}

\def\1{\bm{1}}

\DeclareMathAlphabet{\mathsfit}{\encodingdefault}{\sfdefault}{m}{sl}
\SetMathAlphabet{\mathsfit}{bold}{\encodingdefault}{\sfdefault}{bx}{n}

\DeclareMathOperator*{\argmax}{arg\,max}
\DeclareMathOperator*{\argmin}{arg\,min}

\usepackage{hyperref}
\usepackage{url}
\usepackage{amsmath}
\usepackage{amssymb}
\usepackage{mathtools}
\usepackage{amsthm}

\usepackage{microtype}
\usepackage{graphicx}
\usepackage{subfigure}
\usepackage{url}            
\usepackage{booktabs}       
\usepackage{amsfonts}       
\usepackage{nicefrac}       
\usepackage{microtype}      
\usepackage{tabularx}
\usepackage{microtype}
\usepackage{graphicx}
\usepackage{subfigure}
\usepackage{enumitem}
\usepackage{adjustbox}
\usepackage{mathabx}
\usepackage{booktabs} 
\usepackage{graphicx,latexsym,amsmath,amssymb,amsthm,eucal,graphicx,color}
\usepackage{lipsum}
\usepackage{siunitx}
\usepackage{xcolor}
\usepackage{amsmath}
\usepackage{adjustbox}

\usepackage{rotating}

\usepackage{wrapfig}

\usepackage{kantlipsum,cuted}
\usepackage{tikz}
\usepackage{amssymb}
\usepackage{pifont}
\newcommand{\cmark}{\ding{51}}%
\newcommand{\xmark}{\ding{55}}%

\usepackage[capitalize,compress]{cleveref}
\theoremstyle{plain}
\newtheorem{theorem}{Theorem}[section]
\newtheorem{proposition}[theorem]{Proposition}

\theoremstyle{definition}

\theoremstyle{remark}

\title{Toward a Geometrical Understanding of Self-supervised \\ Contrastive Learning}

\author{\name Romain Cosentino \email rom.cosentino@gmail.com \\
      \addr Ming Hsieh Department of Electrical and Computer Engineering \\
      University of Southern California
      \AND
      \name Anirvan Sengupta  \email anirvans.physics@gmail.com \\
      \addr Flatiron Institute \& 
Rutgers University
      \AND
      \name Salman Avestimehr \email avestime@usc.edu \\
       \addr Ming Hsieh Department of Electrical and Computer Engineering \\
      University of Southern California
       \AND
       Mahdi Soltanolkotabi \email msoltoon@gmail.com \\
        \addr Ming Hsieh Department of Electrical and Computer Engineering \\
      University of Southern California
       \AND
    Antonio Ortega \email aortega@usc.edu \\
\addr Ming Hsieh Department of Electrical and Computer Engineering \\
      University of Southern California
       \AND
       Ted Willke \email ted.willke@intel.com \\
       \addr Intel Labs \\
       \AND 
       Mariano Tepper \email mariano.tepper@intel.com \\
       \addr Intel Labs 
    }

\begin{document}

\maketitle

\begin{abstract}
Self-supervised learning (SSL) is currently one of the premier techniques to create data representations that are actionable for transfer learning in the absence of human annotations. Despite their success, the underlying geometry of these representations remains elusive, which obfuscates the quest for more robust, trustworthy, and interpretable models. In particular, mainstream SSL techniques rely on a specific deep neural network architecture with two cascaded neural networks: the encoder and the projector. When used for transfer learning, the projector is discarded since empirical results show that its representation generalizes more poorly than the encoder's. In this paper, we investigate 
the representation induced by the encoder and how the strength of the data augmentation policies as well as the width and depth of the projector affect its representation.
We discover a non-trivial relation between the encoder, the projector, and the data augmentation strength: with increasingly larger augmentation policies, the projector, rather than the encoder, is more strongly driven to become invariant to the augmentations. It does so by eliminating crucial information about the data by learning to project it into a low-dimensional space, a noisy estimate of the data manifold tangent plane in the encoder representation. This analysis is substantiated through a geometrical perspective with theoretical and empirical results. 
\end{abstract}

\section{Introduction}
\begin{figure}[t]
    \centering
    \includegraphics[width=0.45\linewidth]{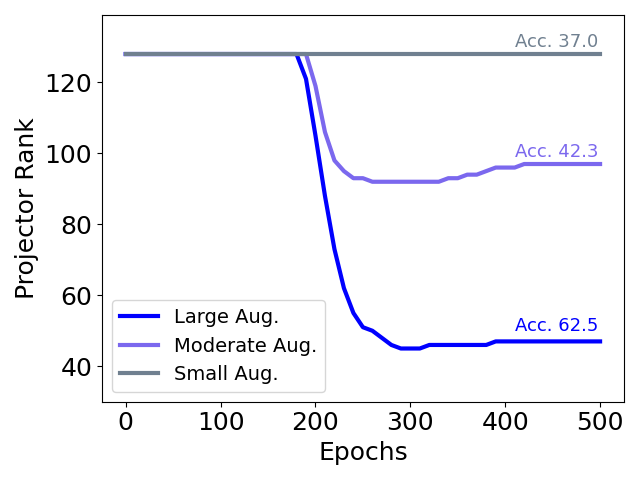}
\vspace{-.1cm}
    \caption{\textbf{Evolution throughout contrastive SSL training of the rank of a linear projector of dimension $512 \times 128$ for different augmentation strengths, and the associated accuracy obtained on Cifar100 by using the representation extracted in the encoder space}. Large, moderate, and small augmentations refer to the strength of the data augmentation applied to the input samples (see Table~\ref{table:aug} for each configuration). The smaller the strength of the data augmentation policy, the less the projector suffers from dimensional collapse. However, when the projector is affected by a substantial dimensional collapse, the encoder representation becomes suitable for the downstream task. In this work, we demystify this intriguing relationship between augmentation strengths, encoder embedding, and projector geometry. 
    }
    \label{fig:rank_augs}
\end{figure}
Training of models that are capable of extracting meaningful data embedding without relying on labels has recently reached new heights with the substantial development of self-supervised learning (SSL) methods. These approaches 
replace labels required for supervised learning with augmentation policies, 
which will define the desired invariances of the trained representation. From these augmentation policies, multiple transformed instances of the same sample are generated, and the network is trained so that their embeddings coincide. After training, the network can be used as a mapping for other datasets to obtain an efficient data representation for various downstream tasks. Most of the SSL algorithms developed have shown competitive performances with supervised learning methods \cite{DBLP:journals/corr/abs-2104-14294, chen2020big, DBLP:journals/corr/abs-2002-05709, DBLP:journals/corr/abs-2006-07733, DBLP:journals/corr/abs-1912-01991, DBLP:journals/corr/abs-2104-14548, DBLP:journals/corr/abs-2105-04906, DBLP:journals/corr/abs-2103-03230, DBLP:journals/corr/abs-1912-01991,DBLP:journals/corr/abs-1807-05520,DBLP:journals/corr/abs-2005-04966,DBLP:journals/corr/abs-2005-10243, DBLP:journals/corr/abs-2006-07733,DBLP:journals/corr/abs-2104-14294}, and more importantly, are more efficient to perform transfer learning for most data distributions and tasks \cite{DBLP:journals/corr/abs-2011-13377,DBLP:journals/corr/abs-2104-14294}. 

While the generalization capability of SSL models has been observed empirically, their inner functioning and the key components to their success remain elusive. Recently, researchers have attempted to capture an understanding of SSL paradigm; In \cite{haochen2021provable}, they provide a graph formulation of contrastive loss functions and leverage it to provide generalization guarantees.  The authors in \cite{DBLP:journals/corr/abs-2110-09348} analyzes contrastive loss functions and the dimensional collapse problem based on the gradient of a linear network. Similarly,\cite{DBLP:journals/corr/abs-2005-10242} analyzes theoretically an end-to-end network mapping the input to output living on the unit sphere and fed into the constrastive loss function. In \cite{huang2021towards} they consider a distance-based theoretical approach to highlight the importance of the data augmentations in constastive SSL. Another line of work, \cite{wang2021understanding}, considers the impact of the temperature parameter used in SSL loss function and how it impacts the data embedding. Another approach considers non-contrastive loss function and performs a spectral analysis of DNN's mapping \cite{tian2021understanding}.

Interestingly, most SSL frameworks developed so far do not use the data embedding provided by the entire trained network's output and instead use the representation extracted from an internal layer. In particular, the deep neural network (DNN) that is used for SSL is usually composed of two cascaded neural networks: the encoder (backbone) and the projector. Usually, the encoder is a residual network, or more recently, a vision transformer, and the projector is an MLP. While the projector output is used for training, only the encoder representation is used for downstream tasks. It is known that the representation at the output of the projector, which is designed to be almost invariant to the selected augmentations, discards crucial features for downstream tasks, such as color, rotation, and shape, while the one at the output of the encoder still contains these features \cite{DBLP:journals/corr/abs-2002-05709,appalaraju2020good}. 

While the work developed so far aiming at understand the behaviour of SSL provide insights into its various aspects, none of them describe the embedding that is used for downstream task: the encoder's output. Their theoretical modelisation of SSL paradigm consider the output of the projector. To open the door to the understanding of the transfer learning capability of DNN trained in a SSL regime, it is crucial to take into account this structural specificity. In this work, we propose to capture \emph{the behaviour of the encoder's network} from a geometrical point of view. In particular our approach aims at demystifying the observation depicted in Fig.~\ref{fig:rank_augs}, where we display the evolution of the rank of weights of a linear projector during training. We observe that there is an interplay between the \emph{generalization capability} of the encoder's representation, the \emph{strength} of the augmentations, and the \emph{rank collapse} of the weights of the projector. In our study, we take a geometrical approach to understand this relationship as well as the benefits on the encoder's representation of using a non-linear projector. Our contribution can be summarized as follows:
\begin{itemize}[leftmargin=*]
  \setlength{\itemsep}{0pt}
  \item \textbf{Interpretable InfoNCE upper bound.} We first derive an interpretable upper bound of the InfoNCE loss function \cite{oord2019representation}, a vastly used loss in SSL. This upper bound is then leveraged throughout the paper to derive insights into the geometry of SSL algorithms (Sec.~\ref{sec:upper_bound}).
  \item \textbf{Dimensional collapse of the projector and data augmentation strength.} We show that the dimensional collapse of the projector depends on $(i)$ the per-sample augmentation strength, $(ii)$  the number of augmentations. We also show how they affect the distribution of the data in the encoder's output space. To do so, we theoretically analyze two forces driving the InfoNCE loss function: undoing the effect of the data augmentation policies and reducing the similarity between each datum and its negative samples (Sec.~\ref{sec:proj_aug_strength}).  
  \item \textbf{Impact of data augmentation strength, depth and width of projector onto encoder's representation.} We show that there exists an intricate relationship between the estimation of the data manifold tangent space performed by the encoder and the strength of the augmentations. In particular, $(i)$ in the case of large augmentations, the estimate of the data manifold tangent space at initialization is poor, but, as training progresses augmentations become beneficial for representing data manifold directions more accurately, $(ii)$ in the case of small augmentations the estimation of the data manifold tangent plane remains poor throughout the training. We also show that with a linear projector, the encoder is constrained to project the data manifold onto a linear subspace, while a non-linear projector enables the encoder to map the tangent space of the data manifold onto a continuous and piecewise affine subspace, therefore unlocking its expressive capability (Sec.~\ref{sec:proj_subspace}).
 \end{itemize}

\section{Background and notations}

\begin{wraptable}{r}{0.5\textwidth}
\vspace{-.9cm}
\centering
\renewcommand{\arraystretch}{1.1}
\caption{Notation reference card}
\label{table:notations}
\centering
\scalebox{.90}{
\begin{tabular}{l l}
\hline
$\boldsymbol{x_i} \in \mathbb{R}^d$ & input data  \\
$\mathcal{T}$ & augmentation distribution \\
$\boldsymbol{x_i^{(k)}} \in \mathbb{R}^d, k=1,2$ & augmented pair \\
$h: \mathbb{R}^{d} \rightarrow \mathbb{R}^{d_{\mathrm{enc}}}$ & encoder mapping\\
$g:\mathbb{R}^{d_{\mathrm{enc}}} \rightarrow \mathbb{R}^{d_{\mathrm{proj}}}$ & projector mapping\\
$f = g \circ h:\mathbb{R}^{d} \rightarrow \mathbb{R}^{d_{\mathrm{proj}}}$ & encoder and projector mapping \\
$\boldsymbol{f_i^{(k)}} = \boldsymbol{g_i^{(k)}} \circ \boldsymbol{h_i^{(k)}}$ & $f(\boldsymbol{x_i^{(k)}}) = g(\boldsymbol{x_i^{(k)}}) \circ h(\boldsymbol{x_i^{(k)}})$ \\
$W$ & linear projector's weights \\
$B_{-i}$ & $ \left \{ (j,k) | j \neq i, k=1,2 \right \}$ \\
$\left \{ \boldsymbol{\tilde{f}_l}, l \in B_{-i} \right \}$ & $ \left \{ \boldsymbol{f_j^{(k)}}| j\neq i, k=1,2 \right \}$ \\
 $\boldsymbol{\tilde{h}^{\star}_i}$ &   $\argmax_{\boldsymbol{\tilde{h}_l}, l \in B_{-i}}  \boldsymbol{h_i^{(1)^T}} WW^T \boldsymbol{\tilde{h}_l}$ \\
 $\boldsymbol{\delta  h_i}$ & $\boldsymbol{h_i^{(2)}} - \boldsymbol{h_i^{\star}}$ \\
 $S$ & $\mathrm{Span}\left \{  \delta \boldsymbol{h_i}, i=1,\dots,N \right \}$\\
 $\mathrm{Proj}(\boldsymbol{h^{(1)}_i})$ & $WW^T \boldsymbol{h^{(1)}_i}$ \\
 
\hline 
\end{tabular}}
\end{wraptable}
\textbf{Notations.}
We will denote by $(\boldsymbol{x_i})_{i=1}^{N}$, the original data in $\mathbb{R}^{d}$, $\boldsymbol{x_i}^{(1)} = t_1(\boldsymbol{x_i}), t_1 \sim \mathcal{T}$ and $\boldsymbol{x_i}^{(2)} = t_2(\boldsymbol{x_i}), t_2 \sim \mathcal{T}$ the augmented pairs obtained by sampling a distribution of augmentation, $\mathcal{T}$, and applying the sampled transformation onto each input data. These augmented pairs are fed into the network $f = g \circ h $ where $h$ is the encoder with output dimension $d_{\mathrm{enc}}$ and $g$ the projector with output dimension $d_{\mathrm{proj}}$. Notations are summarized in Table~\ref{table:notations}.

\textbf{Architecture and hyperparameters.}
For all the experiments performed in this work, we used the SimCLR framework \cite{DBLP:journals/corr/abs-2002-05709}, trained using the InfoNCE loss function. The encoder is defined as a Resnet$18$ with output dimension  $d_{\mathrm{enc}} = 512$. The projector is linear with output dimension $d_{\mathrm{proj}} = 128$. The optimization of the loss function is performed using LARS \cite{you2017large} with learning rate $4.0$, weight decay $10^{-6}$, and momentum $0.9$. The dataset under consideration is Cifar$100$ \cite{krizhevsky2009learning}, sensible tradeoff between the computational resources needed for training and the challenge it poses to Resnet$18$ (i.e., to avoid fitting it with too much ease).

\begin{wraptable}{r}{0.5\textwidth}
\renewcommand{\arraystretch}{0.8}
\caption{Data Augmentation Policies}
\centering
\scalebox{.90}{
\begin{tabular}{l c c c}
\hline
& \multicolumn{3}{c}{Augmentation strength} \\
\cline{2-4}
& Large & Moderate & Small \\
\hline
Horiz.~flipping  & \cmark & \cmark & \xmark  \\
Grayscaling ($p=0.2$) & \cmark & \cmark & \cmark \\
Resized crop  & $0.08 - 1 $ & $0.5 - 1 $ & $0.9 - 1 $\\
Color jittering ($p=0.8$) & $s=0.5$ & $s=0.25$ & $s=0.05$ \\
\hline 
\label{table:aug}
\end{tabular}}
\end{wraptable}

\textbf{Data augmentation policies.}
The augmentation policies to train SimCLR as well as to perform our analysis will be based on: random horizontal flipping, random resized crop, random color jittering, and random grayscale (see \url{https://pytorch.org/vision/stable/transforms.html} for details). Three settings will be tested: small augmentations, moderate augmentations, and large augmentations. The details of the hyperparameters are shown in Table~\ref{table:aug}, where for resized crop, we display the scale parameter, and for color jittering the parameter $s$ that such that brightness $(0.8s)$, contrast $(0.8s)$, saturation $(0.8s)$, and hue $(0.2s)$. Note that the large augmentation setting corresponds to the optimal parameters to obtain the best representation on Cifar$100$.

\begin{figure}[t]
    \centering
    \includegraphics[width=.45\linewidth]{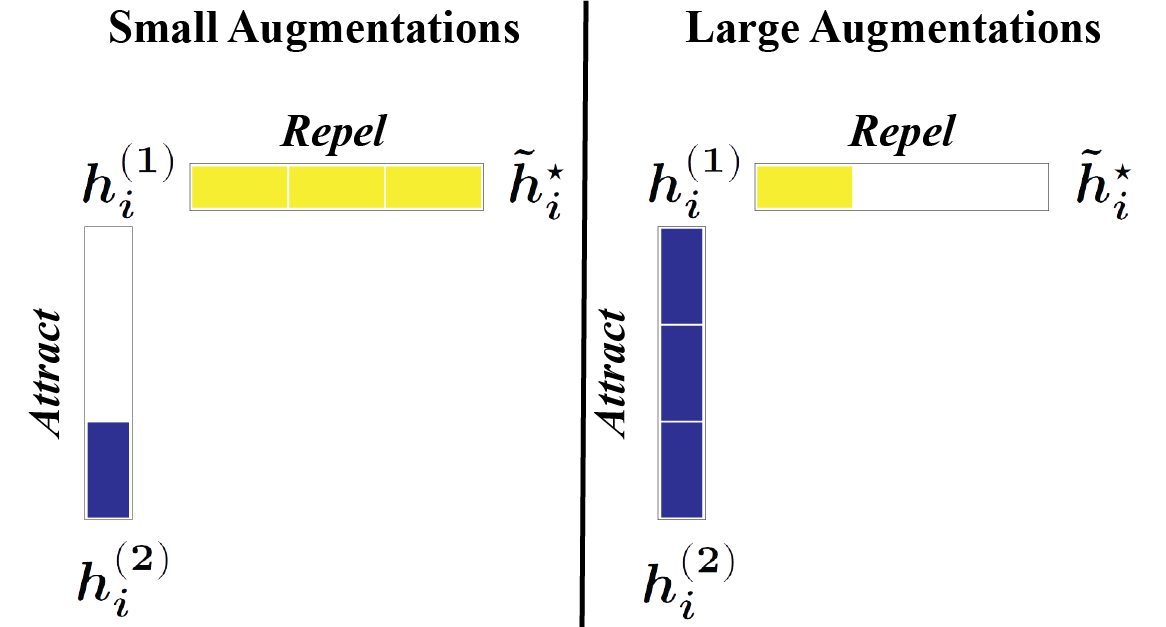}
    \caption{\textbf{The strength of the augmentations (depicted as ``power bars'') acts as a weighting term balancing the attraction of augmented samples and the repulsion effect of the negative samples.} Contrastive SSL introduces negative samples to avoid feature collapse, which aims at enforcing the representation learned to fill the available dimensions. Under the InfoNCE loss, \cref{eq:infoNCE}, there is a trade-off between the repulsion effect induced by the negative samples and the invariance property enforced by maximizing the similarity between two augmented versions of the same sample. We show that the strength of the data augmentation policies acts as a per-datum weight reinforcing the invariance term and reducing the effect of the negative samples. On the contrary, reducing the strength of the augmentations accentuates the repulsing effect from the negative samples. (Notations detailed in Table~\ref{table:notations}).}
    \label{fig:repell_vs_aug}
\end{figure}

\textbf{Loss function.}
The infoNCE loss function, widely used in contrastive learning  \cite{chen2020simple,chen2020big,misra2019selfsupervised,he2020momentum,dwibedi2021little,yeh2021decoupled} is defined as
\begin{align}
\label{eq:infoNCE}
\mathcal{L}_{\mathrm{infoNCE}}&  = - \frac{1}{N} \sum_{i=1}^{N}  \log( \frac{\exp( \beta \boldsymbol{f_i^{(1)^T}} \boldsymbol{f_i^{(2)}}) }{\sum\limits_{l \in B_{-i}} \exp(\beta \boldsymbol{f_i^{(1)^T}} \boldsymbol{\tilde{f}_l}) }) ,
\end{align}
where $\left \{\boldsymbol{\tilde{f}_l}: l \in B_{-i} \right \} = \left \{ \boldsymbol{f_j^{(k)}}: j\neq i, k=1,2 \right \}$. Thus, for each $i$, $B_{-i}$ consists of the indices of the negative samples related to $\boldsymbol{x_i}$, i.e., all the data points except the two augmentations $\boldsymbol{x_i}^{(1)}$ and $\boldsymbol{x_i}^{(2)}$. Note that all  projector outputs are normalized, that is, $||\boldsymbol{f_i^{(k)}} ||_2= 1$.

The numerator of the loss function in \cref{eq:infoNCE} favors a similar representation for two augmented versions of the same data, while the denominator tries to increase the distance between each first augmented sample and the components of all other pairs.
Note that, in practice, this loss is often symmetrized. For all the experiments in this work, the inverse temperature parameter is set to $\beta=2$. 

\textbf{Rank estimation.}
The numerical estimation of the rank is based on the total variance explained as recently used to analyze the rank of DNNs embedding \citet{NEURIPS2020_d5ade38a}. For a given matrix $W$ with singular values $\sigma_1,\dots, \sigma_d$ its estimated rank w.r.t to $\tau$ is defined as
\vspace{-.3cm}
\begin{equation}
\label{eq:rank}
    \mathrm{Rank}_{\tau}(W) = \sum_{j=1}^{d} 1_{ \left \{ |\sigma_j | \geq \tau  \right \} },
\end{equation}
that is, it is the number of singular values whose absolute values are greater than $\tau$.

\section{An interpretable InfoNCE upper bound}
\label{sec:upper_bound}
In this section, we propose an upper bound on the InfoNCE loss that will allow us to derive insights into $(i)$ the impact of the strength of the data augmentation policies on the projector and encoder embeddings and $(ii)$ the nature of the information discarded by the projector and $(iii)$ how does increasing the depth and width of the projector affect the encoder embedding. We provide an overview of InfoNCE in Appendix~\ref{app:infoNCE}. The following proposition illustrates the InfoNCE upper bound in the case of a linear projector.

\begin{proposition}
\label{prop:upperbound_linear}
Considering a linear projector $g(\boldsymbol{h_i^{(k)}}) = W^T \boldsymbol{h_i^{(k)}}, W \in \mathbb{R}^{ d_{enc} \times d_{proj}}$, InfoNCE is upper bounded by
\begin{equation}
\label{eq:upper}
\mathcal{L}_{\mathrm{Upper}} = \beta \mathcal{L}_{\text{invariance}}  + \beta \mathcal{L}_{\text{repulsion}} + \log(2(N-1)),
\end{equation}
where
\vspace{-.6cm}
\begin{align}
    \mathcal{L}_{\text{invariance}} &= - \frac{1}{N}\sum_{i=1}^{N} \operatorname{Sim}(W^T \boldsymbol{h_i^{(1)}}, W^T \boldsymbol{h_i^{(2)}}) ,
    \label{eq:invariance}
    \\
    \mathcal{L}_{\text{repulsion}} &= \frac{1}{N}\sum_{i=1}^{N} \operatorname{Sim}(W^T \boldsymbol{h_i^{(1)}}, W^T \boldsymbol{\tilde{h}^{\star}_i}) ,
    \label{eq:repulsion}
\end{align}
\vspace{-.1cm}
$\operatorname{Sim}(\boldsymbol{x},\boldsymbol{y})= \boldsymbol{x^T} \boldsymbol{y} / (\left \| \boldsymbol{x} \right \| \left \| \boldsymbol{y} \right \|)$ and $\displaystyle \boldsymbol{\tilde{h}^{\star}_i} =  \argmax_{\boldsymbol{\tilde{h}_l}, l \in B_{-i}} \operatorname{Sim}(W^T \boldsymbol{h_i^{(1)^T}}, W^T \boldsymbol{\tilde{h}_l})$. (Proof in Appendix~\ref{proof:upperbound_linear}).
\end{proposition}

\begin{wrapfigure}{r}{0.5\textwidth}
\vspace{-.5cm}
    \centering
    \includegraphics[width=1\linewidth]{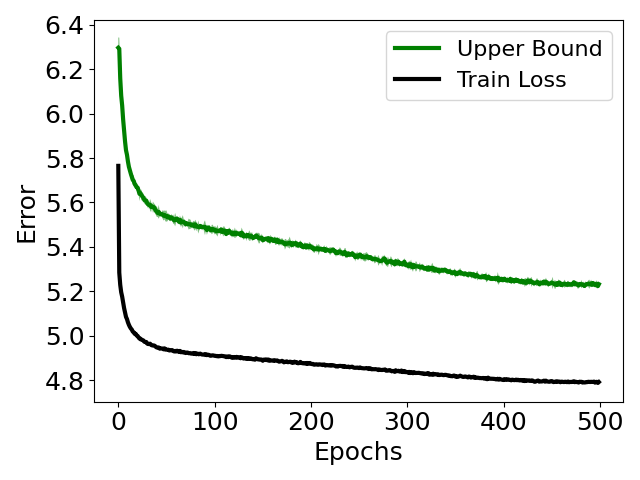}
    \vspace{-.8cm}
    \caption{\textbf{The upper bound $\mathcal{L}_{\mathrm{Upper}}$ in \cref{eq:upper} tracks the InfoNCE loss} (mean and standard deviation over $5$ runs) throughout training with a batch size of $128$.
    }
    \label{fig:upper_bound}
\end{wrapfigure}

While in the InfoNCE for each augmented sample the negative samples taken into account are within a ball of radius defined by $\beta$ (Appendix~\ref{app:infoNCE} for detailed explanations), in our upper bound in \cref{eq:upper}, we only consider the negative sample resembling the most with its corresponding first augmented sample, denoted by $\tilde{h}_i^{\star}$.

The bound $\mathcal{L}_{\mathrm{Upper}}$ of \cref{eq:upper} captures the essential trade-offs in SSL: $\mathcal{L}_{\text{invariance}}$ corresponds to the maximization of the similarity between two versions of the same augmented data in the projector's output space, 
while $\mathcal{L}_{\text{repulsion}}$ approximates the repulsion term that aims at suppressing the collapse of the representation by using the closest negative sample. 
In \cref{fig:upper_bound}, we show that InfoNCE and $\mathcal{L}_{\mathrm{Upper}}$ behave similarly during training. While tightening the bound seems possible, $\mathcal{L}_{\mathrm{Upper}}$ is intuitive and sufficient to support our analysis.

\section{Dimensional collapse hinges on augmentations strength}
\label{sec:proj_aug_strength}
In this section, we leverage the upper bound in \cref{eq:upper} to show that the dimensional collapse of the projector network is directly related to the strength of the augmentation policies as well as the number of augmentations, as observed in Fig.~\ref{fig:rank_augs}. In particular, the stronger the augmentations, the lower the rank of the projector weights. We also show how the augmentations affect the distribution of the data in the encoder's embedding. We first consider 
$\mathcal{L}_{\text{invariance}}$ and $\mathcal{L}_{\text{repulsion}}$ from \cref{eq:upper} separately, before addressing their interaction. We provide in Fig.~\ref{fig:repell_vs_aug} a description of the interactions between the invariance term and repelling term with respect to the strength of the augmentations.

\textbf{(1)} 
The \textbf{invariance term} $\mathcal{L}_{\text{invariance}}$ in \cref{eq:invariance} promotes the augmented samples to coincide; we show here how it affects the rank of the projector weights and how stronger augmentations cause the projector mapping to be invariant. We first develop some intuitions by modeling the transformations as a linear action in the encoder space. While this approach can always be achieved as the output of the encoder defines a vector space, we will then show that modeling the augmentations as Lie group transformations will provide more insights.

\begin{figure}[t]
    \centering
    \includegraphics[width=0.45\linewidth]{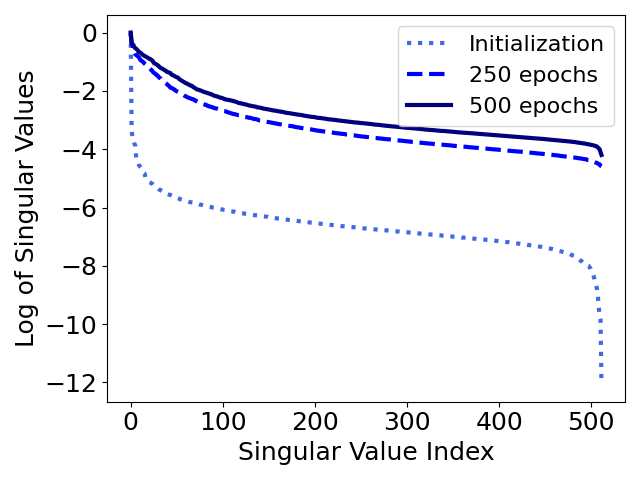}
        \vspace{-.1cm}
    \caption{\textbf{Encoder space log-singular values} for Cifar$100$ dataset ($d_{\text{enc}} = 512$). The log-spectrum is evaluated at different training time: initialization (\textit{dotted line}), half-training time (\textit{dashed line}), and after training (\textit{solid line}). In these three settings, the estimated rank (see \cref{eq:rank}) leads to $\mathrm{Rank}[\boldsymbol{h_i^{(k)}}, \dots, \boldsymbol{h_i^{(k)}}] \approx 512$, i.e., the embedded transformed samples span all the available directions in the encoder space (albeit a slight rank deficiency at initialization).}
    \label{fig:rank_enc_cifar_linear}
\end{figure}

\textit{(a) Linear transformations in encoder space.} This additive relationship between augmented pair in the encoder space can be formally described by $\forall i, \boldsymbol{h_i^{(2)}} = \boldsymbol{h_i^{(1)}} + \boldsymbol{v_i}$, where $\boldsymbol{v_i} \in \mathbb{R}^{d_{\mathrm{enc}}}$. Note that this formalism is always verified in practice. Following this approach, the invariance term can be re-written as
\vspace{-.15cm}
\small{
\begin{align}
     \mathcal{L}_{\text{invariance}} & = \frac{1}{N} \sum_{i=1}^{N} - \mathrm{Sim}(W^T \boldsymbol{h_i^{(1)}}, W^T (\boldsymbol{h_i^{(1)}} + \boldsymbol{v_i}  ) ), 
     \label{eq:linear}
\end{align}}
\normalsize
which can be minimized by maximizing the cosine similarity between $W^T \boldsymbol{h_i^{(1)}}$ and $W^T (\boldsymbol{h_i^{(1)}} + \boldsymbol{v_i})$. This can be achieved by projector weights, i.e., $W$, such that: $(i)$ $W^T \boldsymbol{v_i}$ is colinear to $W^T \boldsymbol{h_i^{(1)}}$ or $(ii)$ the $\boldsymbol{v_i}$ are in the null space of $W$. In both cases, the cosine similarity is maximized. This observation leads to the following proposition.

\begin{proposition}
\label{prop:kernel_linear}
Assuming $\boldsymbol{h}_i^{(1)}$ and $\boldsymbol{h}_i^{(2)}$ are not colinear for all $i=1,\dots,N$, then $\mathcal{L}_{\text{invariance}}$ is minimized if and only if $v_i \in \mathrm{Ker}(W)$. That is, the direction of augmentations in the encoder space belong to the kernel of the projector's weights. (Proof in Appendix \ref{proof:kernel_linear})
\end{proposition}

Therefore, by the rank-nullity theorem, the rank of the projector weights decreases as the dimension of the span of the direction of the augmentations increases, i.e., as $\mathrm{Rank}(\boldsymbol{v_1},\dots,\boldsymbol{v_N})$ becomes large. Interestingly, if all these augmented directions are mapped onto a low-dimensional subspace by the backbone encoder, then the rank of the projector weights can be maintained which is formalized by the following propositon.

\begin{proposition}
\label{prop:rank_nullity}
The lower the rank of the subspace the encoder maps the augmented directions onto, the lower achievable dimensional collapse of the projector. (Proof in Appendix~\ref{app:rank_nullity}).  
\end{proposition}

Intuitively, the dimension of the span of the $\boldsymbol{v_i}$'s increases if augmentations point in different directions in the encoder space. Besides, the larger the spread (i.e., the difference between the maximum and minimum strength) of the augmentation policy distribution $\mathcal{T}$, the higher the dimension of the span of the $\boldsymbol{v_i}$. We provide empirical evidence regarding this aspect in Appendix~\ref{app:rank_vs_rot_toy}.

\textit{(b) Non-linear transformations in encoder space.}
We formalize non-linear transformations as the result of the application of Lie groups onto the data embedded in the encoder space. This is motivated by the fact that the encoder output is a continuous piecewise affine low dimensional manifold \cite{DBLP:journals/corr/abs-1905-12784,balestriero2019geometry}, in which the Lie group defines transformations between points that can model those observed in various natural datasets \cite{connor2021variational,pmlr-v145-cosentino22a}. 

Formally, $\boldsymbol{h_i^{(2)}} = \exp( \epsilon_i G) \boldsymbol{h_i^{(1)}}$, where $G \in \mathbb{R}^{d_{\mathrm{enc}} \times d_{\mathrm{enc}}}$ is the generator of a Lie group that characterizes the type of transformation and $\epsilon_i$ is a scalar denoting the strength of the transformations induced by $G$ applied onto the $i^{th}$ datum. A primer on Lie group transformations is provided in Appendix~\ref{app:liegroup}. Note that in practice, the $\epsilon_i$ are the strength parameters that are sampled from the policy distribution and the type of transformation is defined by the selected augmentation policy, e.g., crop, colorjitter, horizontal flipping. 

\begin{figure}[t]
    \centering
    \includegraphics[width=0.45\linewidth]{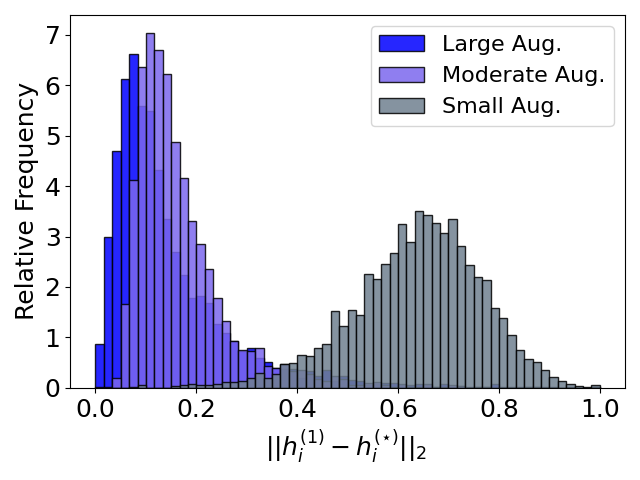}
        \vspace{-.1cm}
    \caption{\textbf{Visualization of the normalized Euclidean distance between $\boldsymbol{h_i^{(1)}}$ and $\boldsymbol{h_i^{\star}}$} for different training configurations: small, moderate, and large augmentations. Under large augmentations, the repulsion term $\mathcal{L}_{\text{repulsion}}$ is deemphasized. Under small augmentations, $\mathcal{L}_{\text{repulsion}}$ dominates the loss. The augmentation strength act as a per-datum scalar weight that governs how much the representation needs to be invariant as opposed to how much each augmented sample and their nearest neighbor should be projected onto opposite directions.}
    \label{fig:h1_hstar_distance}
\end{figure}

We can now express $\mathcal{L}_{\text{invariance}}$ as a function of the strength parameter as well as the generator of the transformation.
\small{
\begin{align}
     \mathcal{L}_{\text{invariance}} & = -\frac{1}{N} \sum_{i=1}^{N} \mathrm{Sim}(W^T \boldsymbol{h_i^{(1)}}, W^T \exp( \epsilon_i G) \boldsymbol{h_i^{(1)}} ).
\label{eq:liegroup}
\end{align}}
\normalsize
We observe that, $(i)$ if the sampled strength $\epsilon_i$ is $0$, then the similarity is maximized and therefore the invariance term does not penalize the loss function, and $(ii)$ the larger the strength parameter is the more $\exp( \epsilon_i G) \boldsymbol{h_i^{(1)}}$ differs from $\boldsymbol{h_i^{(1)}}$ thus penalizing the loss function, except if the generator  of the transformation, i.e., $G$, is in the kernel of $W$ which we now discuss. Therefore, the strength of the augmentation acts as a balancing term between the invariance loss and the repulsion loss. In particular, the strength of the transformation corresponds to a per-datum scalar weight that governs how much the transformation affects the projector mapping toward an invariant representation, as depicted in Fig.~\ref{fig:repell_vs_aug}.  

The following proposition provides insights into the dimensional collapse of the projector and its relationship with the augmentation policy. 

\begin{proposition}
\label{prop:kernel_nonlinear}
Assuming $\boldsymbol{h_i}^{(1)}$ and $\boldsymbol{h_i}^{(2)}$ are not colinear for all $i=1,\dots,N$ then $\mathcal{L}_{\text{invariance}}$ is minimized if and only if $G \in \mathrm{Ker}(W)$. That is, the generator of the augmentation policy belongs to the kernel of the projector's weights. (Proof in Appendix~\ref{app:kernel_nonlinear}).
\end{proposition}

This proposition shows that, in order to maximize the similarity, the projector weights, i.e., $W$, have to align its kernel with the generators underlying the augmentation policies. Note that we consider here only one group generator $G$ while in practice one can be associated to each augmentation type. Thus, increasing the number of transformations will increase the dimension of the null space of $W$, except if their generator spans similar directions.

\begin{figure*}[t]
\begin{tabular}{@{\hspace{0pt}} c @{\hspace{0pt}} c @{\hspace{0pt}} c @{\hspace{0pt}}}
    \begin{small}Small Augmentations\end{small} & \begin{small}Moderate Augmentations\end{small} & \begin{small}Large Augmentations\end{small} \\

    \includegraphics[width=0.33\linewidth]{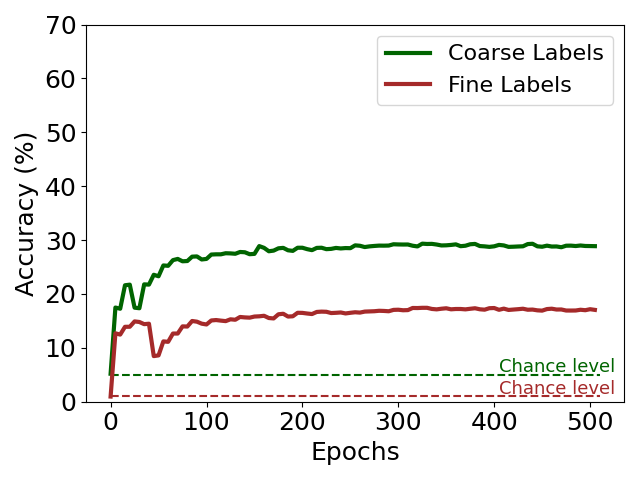}
    &
    \includegraphics[width=0.33\linewidth]{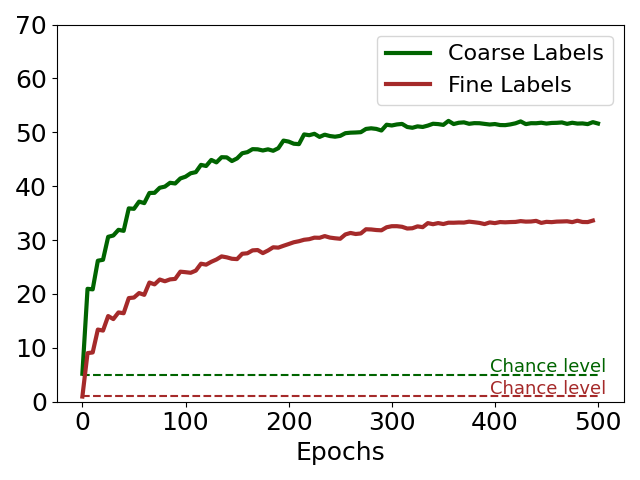}
    &
    \includegraphics[width=0.33\linewidth]{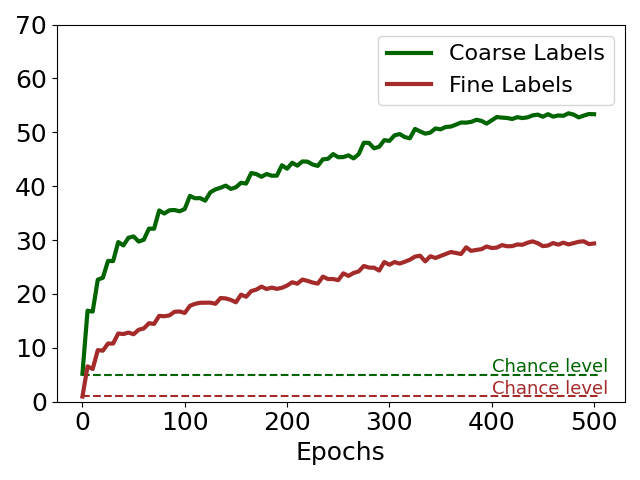}
\end{tabular}
\vspace{-.1cm}
\caption{\textbf{Percentage of matching labels between $\boldsymbol{\tilde{h}_i^{\star}}$ and $\boldsymbol{h_i^{(2)}}$} for the $100$ fine and $20$ coarse labels of the Cifar$100$ dataset when training with different augmentation strengths. Recall that the index of $\boldsymbol{\tilde{h}_i^{\star}}$ is determined in the projector's output space as defined in \cref{prop:upperbound_linear}. We observe that during training, across all augmentation regimes, the amount of shared semantic information increases. As the augmentation strength increases, the training time for the converge of this semantic sharing increases as well. Under small augmentations it converges after a couple hundred epochs, whereas under large augmentations it did not converge yet after $500$ epochs. The strength of the augmentation establishes the trade-off between the invariance and repulsion terms (Sec.~\ref{sec:proj_aug_strength}): under small small augmentations, the repulsion term dominates, which results in a poor representation of the data manifold.}
\label{fig:who_is_h}
\end{figure*}

\textbf{(2)} 
We now consider the \textbf{repulsion term} $\mathcal{L}_{\text{repulsion}}$ in \cref{eq:repulsion}. This term is minimized if for each $i=1,\dots,N$, we have 
$\operatorname{Sim}(W^T \boldsymbol{h_i^{(1)}}, W^T \boldsymbol{\tilde{h}^{\star}_i}) =-1$. That is, the angle between $W^T \boldsymbol{h_i^{(1)}}$ and $W^T \boldsymbol{\tilde{h}^{\star}_i}$ is $\pi$. To do so, the projector maps $\boldsymbol{h_i^{(1)}}$ and its nearest neighbor $\boldsymbol{\tilde{h}_i^{\star}}$ onto diametrically opposed directions. 

If it is feasible for the projector to map each $\boldsymbol{h_i^{(1)}}$ and $\boldsymbol{\tilde{h}_i^{\star}}$, onto diametrically opposed directions, the representation provided would not encapsulate any information regarding the data manifold. It is therefore necessary that the invariance term dominates the loss function to capture information regarding the data manifold. As we mentioned, the strength of the augmentation acts as a per-sample balancing term between the invariance term and the repulsion term. This enables us to understand why current practices in SSL consider the use of large augmentations.

In Fig.~\ref{fig:h1_hstar_distance} we provide the histogram of the distances between $\boldsymbol{h_i^{(1)}}$ and $\boldsymbol{\tilde{h}_i^{\star}}$. As we understood from the previous discussions, when the augmentations are small, this distance tends to be higher as the  $\mathcal{L}_{\text{repulsion}}$ term has a greater impact than $\mathcal{L}_{\text{invariance}}$ on the InfoNCE loss. Conversely, for stronger augmentations, the $\mathcal{L}_{\text{invariance}}$ term dominates.


Note that, in this section, we assume that the linearization capability of the resnet with respect to the transformations enables us to express the transformations with respect to the the generator of the transformation and the per-sample strength. 
In practice, this linearization capability might be limited, and therefore, the projector mapping discards more than just the generator induced by the augmentation policies. This loss of information has been empirically observed in \cite{DBLP:journals/corr/abs-2002-05709,appalaraju2020good}.

\section{Large augmentations and deep projector benefit encoder's estimation of the data manifold}
\label{sec:proj_subspace}
In the previous section we showed the relationship between the dimensional collapse of the projector, the number and strength of augmentations, and the distribution of the data in the encoder's embedding. We now propose to analyze the relationship between the encoder and projector mappings through the lens of data manifold estimation and show how augmentations strength as well as architecture of the projector affect the quality of the encoder embedding.  For sake of clarity, we first develop the case of linear projector, then, we resolve the case of non-linear projector.

We focus our analysis by considering the following reformulation of the upper bound in \cref{eq:upper} with the assumption that the output of the projector are normalized
\begin{align}
     \mathcal{L}_{\mathrm{Upper}} &=  \frac{1}{N} \sum_{i=1}^{N} - \beta \boldsymbol{ \delta h_i}^T  \mathrm{Proj}(\boldsymbol{h_i^{(1)}}) \nonumber \\
    & \hspace{1cm}+ \log(2(N-1)) ,
    \label{eq:upper_sphere_proj}
\end{align}
with  $\boldsymbol{\delta h_i} = \boldsymbol{h_i^{(2)}} -  \boldsymbol{\tilde{h}^{\star}_i}$ and where in the case of a linear projector
\begin{equation}
\mathrm{Proj}(\boldsymbol{h_i^{(1)}}) = WW^T \boldsymbol{h_i^{(1)}},
\end{equation}
i.e., the projection of $\boldsymbol{h_i^{(1)}}$ onto the column space of $W$. 

\begin{figure}[t]
    \centering
    \includegraphics[width=0.45\linewidth]{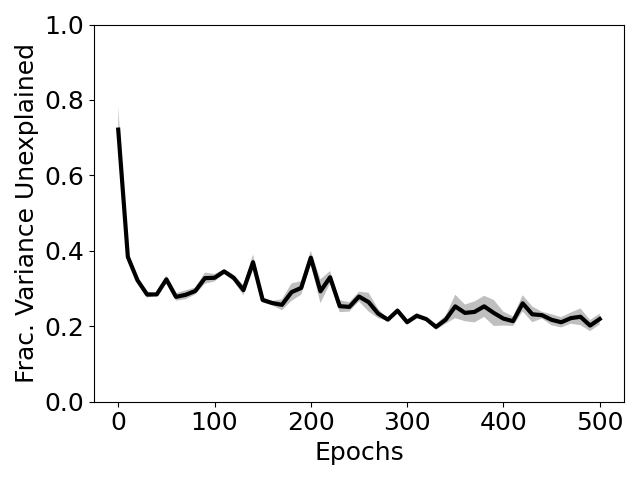}
    \vspace{-.1cm}
    \caption{\textbf{Fraction of variance unexplained, \cref{app:var_unex}, as a measure of misalignment between the column space of the weights of the linear projector, i.e., $W$, and $\mathcal{S}$}, where $\mathcal{S} = \mathrm{Span} \left \{ \boldsymbol{\delta h_i}, i=1,\dots, N \right \}$ defined as (mean and standard deviation over $5$ runs). We observe that the amount of unexplained variance decreases drastically after the initialization, showing that the projector attempts to approximates the subspace $\mathcal{S}$. 
    }
    \label{fig:span_distance}
\end{figure}

To minimize \cref{eq:upper_sphere_proj}, $W$ should enable the projection of $\boldsymbol{h_i^{(1)}}$ onto $\boldsymbol{\delta h_i}$. Therefore, the columns of $W$ needs to include in their span $\mathcal{S} = \mathrm{Span} \left \{ \boldsymbol{\delta h_i}, i=1,\dots, N \right \}$. We empirically verify this behaviour; specifically, in Fig.~\ref{fig:span_distance} we show that during training the fraction of variance of $S$ unexplained by $W$ is decreasing, where the fraction of variance of $S$ unexplained by $W$ is computed as follows
\begin{equation}
    \sum_i \min_{t_i \in \mathbb{R}^{d_{\mathrm{proj}}}} \left \| \boldsymbol{\delta h_i} - W t_i \right \|^2_2 /   \sum_i \left \| \boldsymbol{\delta h_i} \right \|^2_2.
    \label{app:var_unex}
\end{equation}
We posit that the remaining part of the unexplained variance is mainly due to the fact that $(i)$ the encoder embedding of the  $\boldsymbol{\delta h_i}$ is non-linear, $(ii)$ the Resnet$18$ backbone encoder is not capable of linearizing $\mathcal{S}$. We now propose to characterize the subspace $\mathcal{S}$ that appears to be crucial in determining the interplay between the projector and encoder mappings. 

\paragraph{Noisy estimate of the data manifold tangent plane.}
We now consider the $\boldsymbol{\delta h_i}$ as displacement vectors that are the noisy estimate of a manifold tangent plane. This modelization of manifold tangent space has been commonly used in manifold estimation techniques, for more details the reader can refer to \cite{tenenbaum2000global,bengio2005non,bengio2005non2,NIPS2017_86a1793f}. We are interested in knowing how close to the data manifold this estimated tangent space is and how the strength of the augmentation affects this estimation process.
We depict in Fig.~\ref{fig:who_is_h} the amount of label sharing between $\boldsymbol{\tilde{h}_i^{\star}}$ and $\boldsymbol{h_i^{(2)}}$, as training progresses. Note that Cifar$100$ contains $100$ fine and $20$ coarse classes. 
We observe that the semantic similarity between $\boldsymbol{h_i^{(2)}}$ and $\boldsymbol{\tilde{h}_i^{\star}}$ increases with the augmentation strength. Therefore, large augmentations make  $\boldsymbol{\delta h_i}$ to be a close estimate of the data manifold tangent space.

This observation can be understood from the following reasoning: $(i)$ In the large augmentations regime, at initialization, $\boldsymbol{h_i^{(2)}}$ and $ \boldsymbol{\tilde{h}^{\star}_i}$ are distant from each other (as $\boldsymbol{\tilde{h}_i^{\star}}$ is near $\boldsymbol{\tilde{h}_i^{1}}$). Thus, the directions spanned by $\boldsymbol{\delta h_i}$ are a coarse estimation of the underlying data manifold. However, from Sec.~\ref{sec:proj_aug_strength}, we know that during training, $\mathcal{L}_{\text{invariance}}$ dominates, so that $\boldsymbol{h_i^{(2)}}$ becomes closer to $\boldsymbol{h_i^{\star}}$. Therefore, for large augmentations, the estimation of the data manifold tangent plane becomes finer as training progresses. This phenomenon is observed in Fig.~\ref{fig:who_is_h}, where the percentage of $\boldsymbol{\tilde{h}_i^{\star}}$ sharing similar semantic content with $\boldsymbol{h_i^{(2)}}$ steadily increases during training. $(ii)$ In the small augmentations regime, at initialization, $\boldsymbol{h_i^{(2)}}$ and $\boldsymbol{\tilde{h}_i^{\star}}$ are close to each other, but do not share the same label. During training, the dissimilarity between $\boldsymbol{\tilde{h}_i^{\star}}$ and $\boldsymbol{h_i^{(1)}}$ increases as $\mathcal{L}_{\text{repulsion}}$ dominates the loss as discussed in Sec.~\ref{sec:proj_aug_strength} and observed in Fig.~\ref{fig:who_is_h}. The $\mathcal{L}_{\text{repulsion}}$ term of the loss does not favor specific directions. Therefore, it is hard to characterize how (the approximation of) the data manifold tangent plane in the encoder space evolves during training in the case of small augmentations.

\begin{figure}[t!]
    \centering
    \includegraphics[width=.45\linewidth]{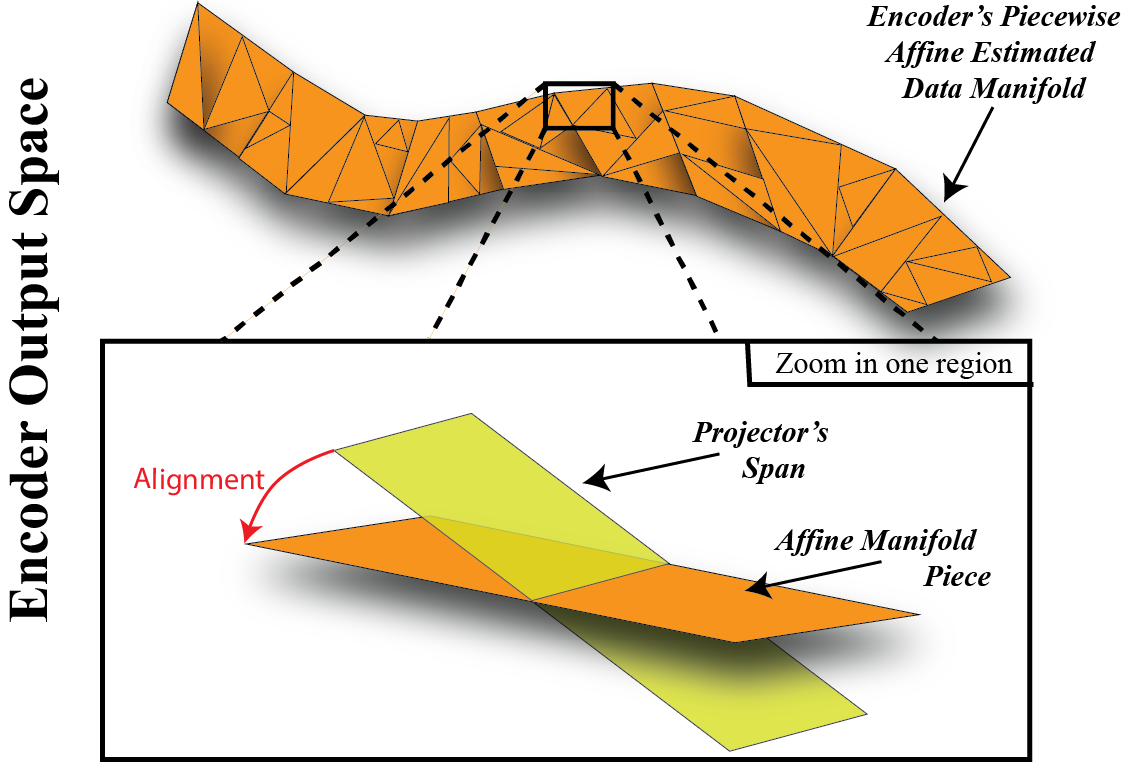}
        \vspace{-.1cm}
    \caption{\textbf{Intuitive illustration of the projector aiming at aligning its column span with the noisy estimate of the data manifold tangent space performed by the encoder.} We provide here in the encoder output space the visualization of the noisy estimate of the data manifold: a continuous and piecewise affine surface representing $\mathcal{S}$ (top). We shed light on the aim of the projector by zooming inside a region of this estimated manifold: it aligns the column span of its mapping (yellow) with the local estimated data manifold tangent space. In the case of a linear projector, the column of $W$ should align with the entire estimated data manifold (top). Therefore forcing the encoder to provide a linear estimate of the data manifold. However, in the case of a non-linear projector, this alignment is performed locally. As the span of the projector is also a continuous and piecewise affine, using a non-linear projector lifts such a constraint on the encoder. The deeper and wider the projector is, the larger is its number of regions, therefore, the less constrained the encoder is with respect to its estimation of the data manifold tangent space. Note that our proposed analysis on the strength of the augmentations can be applied to the non-linear case by considering it locally, i.e., for each local affine map.
     }
        \label{fig:insights}
\end{figure}

Our geometrical understanding of the projector is displayed in Fig.~\ref{fig:insights} and can be summarized as follows: For each batch, one obtains in the encoder space an estimate of the data manifold directions by taking the difference between $\boldsymbol{h_i^{(2)}}$ and $\boldsymbol{\tilde{h}_i^{\star}}$, then, the projector attempts to fit the subspace spanned by these vectors, defined as $\mathcal{S}$. Given the nature of this subspace, we conclude that the projector attempts to align each datum, embedded in the encoder space, onto the directions of the estimated data manifold tangent space. Therefore the projector discards the information that is not part of the subspace $\mathcal{S}$. From this, we understand that in the case of small augmentations, due to the proximity of $\boldsymbol{h_i^{(1)}}$, $\boldsymbol{h_i^{(2)}}$, and $\boldsymbol{\tilde{h}_i^{\star}}$ at initialization, the error performed by the projector is small in order to obtain the desired projections. This also highlights why, in Fig.~\ref{fig:rank_augs}, the rank of the projector weights is not modified when training under small augmentations.

\textbf{The effect of depth and width of non-linear projector.}
We now discuss how the width and depth of a non-linear projector affects its objective of projecting each $\boldsymbol{h_i^{(1)}}$ onto $\boldsymbol{\delta h_i}$. In particular, we show that when using an MLP, the approximation of $\mathcal{S}$ is performed locally, i.e.,  each part of the encoder output space will be mapped by a different (local) affine transformation induced by the projector. Each affine transformation depends on both the type of non-linearity and the weights of the projector's MLP. Therefore, the aforementioned discussions regarding the strength of the augmentations transfer to this case in a local manner.

We consider MLP projectors employing non-linearities such as (leaky-)ReLU, which are continuous piecewise affine operators. The particularity of these operators is that they induce a partition of their input space. In the present case, we consider the partition induced by the MLP projector, that is, of the output space of the encoder mapping. We will denote this partition as $\Omega$ of $\mathbb{R}^{d_{enc}}$. For more details regarding this approach considering the partitioning of DNN in their input space, the reader should refer to \cite{balestriero2018spline,balestriero2019geometry,DBLP:journals/corr/abs-2009-09525}. Following this exact formulation, the projector mapping can now be expressed using the following closed-form (zeros-bias MLP projector is used for the sake of clarity)
\begin{equation}
\label{eq:cpa}
    g(\boldsymbol{h_i^{(k)}}) = \sum_{\omega \in \Omega} 1_{\{ \boldsymbol{h_i^{(k)}} \in \omega\}}\left(W_{\omega}^T \boldsymbol{h_i^{(k)}} \right),
\end{equation}
where $\Omega$ defines a partition of $\mathbb{R}^{d_{\mathrm{enc}}}$, i.e., the encoder's output. For each region $\omega \in \Omega$ in the encoder's output space (as depicted by polytopes on top of Fig.~\ref{fig:insights}), the projector acts as a linear mapping, induced by the weights $W_{\omega}$. 

It is now clear that whereas in the linear case the InfoNCE loss adjusts $\mathrm{Proj}(\boldsymbol{h_i^{(1)}}) = WW^T \boldsymbol{h_i^{(1)}}$ to fit $\boldsymbol{\delta h_i}$, in the non-linear case, this fitting is performed locally; for each region $\omega$ in the encoder space, the corresponding $W_{\omega}$ is trained to fit the $\boldsymbol{h_i^{(1)}}$ samples belonging to the region $\omega$ onto the subspace $\mathcal{S}$. This is formally described as $\mathrm{Proj}(\boldsymbol{h_i^{(1)}}) = \left \{W_{\omega} W_{\omega}^T \boldsymbol{h_i^{(1)}}| \boldsymbol{h_i^{(1)}} \in \omega \right \} \approx \boldsymbol{\delta h_i}$. 

The deeper and wider the projector MLP is, the larger the number of regions $\omega$ and the smaller their volumes is \cite{montufar2021sharp}. Therefore, in the deep projector regime, the $W_{\omega}$ are trained to projector only one sample, $\boldsymbol{h}_i^{(1)}$, onto one estimate of the tangent space $\boldsymbol{\delta h_i}$. Thus, the granularity of the projection and the fitting of the displacement vector is getting more refined as the depth increases as the number of local mappings increases.

Therefore, while a linear projector forces the encoder to linearize the subspace $\mathcal{S}$, a non-linear projector can map this subspace to a non-linear manifold. The wider and deeper the non-linear projector's MLP is, the less the encoder mapping is constrained. There is therefore a balance between not constraining at all the encoder and enabling it to unlock its expressive power. We now understand that the cross-validation observed regarding the architecture of the projector aimed at finding this balance. Our considerations allows us to understand why in \cite{DBLP:journals/corr/abs-2006-10029}, it was empirically shown that having deeper and larger MLPs could help to improve the accuracy of downstream tasks.

\section{Related Work}

\begin{table}[htbp]
\begin{adjustbox}{width=\columnwidth,center}
\centering
\begin{tabular}{l|cc|cccc}
\hline
{} & \multicolumn{2}{c|}{DNN Mapping Insights} & \multicolumn{4}{c}{{Data Augmentation Insights}}  \\ 
\cline{1-5} \cline{6-7}
{} & {Encoder} & {Projector}  & {Theoretical} & {Empirical} & {Theoretical Aug.} & {Empirical Aug.}  \\ 
{} & {Analysis} & {Analysis}  &  {Analysis}   &  {Analysis} & {Model} & {Model}  \\
\hline
Our work& Yes & Linear$\setminus$Nonlinear   & Encoder & Encoder & Lie generator & Conv. practice   \\
\cite{DBLP:journals/corr/abs-2110-09348} & No & Linear &  Projector  & Encoder &  Additive noise & Gaussian noise  \\ 
  \cite{huang2021towards} &  \multicolumn{2}{c|}{{No separate analysis}} & Projector & Encoder & ($\sigma, \delta)$-augmentation & Conv. practice  \\
\cite{DBLP:journals/corr/abs-2005-10242} &  \multicolumn{2}{c|}{{No separate analysis}}    & Projector & Encoder & - & Conv. practice    \\
 \cite{haochen2021provable} & \multicolumn{2}{c|}{{No separate analysis}}   & Projector & - & Gaussian noise & -  \\  
 \cite{DBLP:journals/corr/abs-2005-10243}&  \multicolumn{2}{c|}{{No separate analysis}}     & Projector & Encoder & - & Conv. practice  \\
 \cite{wang2021understanding} &  \multicolumn{2}{c|}{{No separate analysis}}     & - & Encoder & - & Conv. practice  \\
\hline
\end{tabular}
\label{table:compare}
\end{adjustbox}
\end{table}

We summarize in Table~\ref{table:compare} the main differences with the related work. There are three key aspects differentiating this work. (a) \textbf{Encoder Analysis}: We provide both theoretical and empirical insights on the encoder embedding, i.e., the actual output used for downstream tasks. In contrast, the authors in \cite{DBLP:journals/corr/abs-2110-09348} do not analyze the learning in the encoding network. They investigate an end-to-end linear network that directly connects the input to the overall output fed to the loss function. Similarly,\cite{DBLP:journals/corr/abs-2005-10242} analyzes theoretically an end-to-end network mapping the input to output living on the unit sphere and fed into the loss function. (b) \textbf{Nonlinear Projector}: Our analysis describes how a nonlinear projector impacts the encoder embedding and analyze the effect of its depth and width on the geometry of the encoder. 
(c) \textbf{Lie Group}: While existing approaches \cite{DBLP:journals/corr/abs-2110-09348,huang2021towards,haochen2021provable} are distance-based, our provides a way to capture geometric information via the Lie generator

\section{Conclusions}

In this work, we have investigated the intricate relationship between the strength of the policy augmentations, the encoder representation, and the projector's effective dimensionality in the context of contrastive self-supervised learning. Our analysis justifies the use of large augmentations in practice. Under small augmentations, the data representations are haphazardly mutually repulsed, leading them to reflect the semantics poorly. However, under large augmentations, SSL attempts to approximate the data manifold, extracting in the process useful representation for downstream transfer tasks. However, we also showed that large augmentations forces the projector mapping to discard a large amount of information to project each transformed data onto the estimate of the data manifold. Therefore, discarding the projector mapping for extracting data embedding is necessary as long as SSL requires large augmentations. Pursuit of more interpretable contrastive SSL without projector will require the development of a framework efficient in a small augmentations regime.

\bibliography{tmlr}
\bibliographystyle{tmlr}

\appendix
\section{Appendix}

\section{Lie Group}
\label{app:liegroup}

A Lie group is a group that is a differentiable manifold. For instance, the group of rotation $SO(2) = \left \{ \begin{pmatrix}
 \cos(\theta) & -\sin(\theta)\\ 
  \sin(\theta) & \cos(\theta)  
\end{pmatrix} | \theta \in \mathbb{R}/ 2 \pi \mathbb{Z} \right \}.$

One of the main advantage of having a group with a differentiable manifold structure is that it can be defined by an exponential map: $SO(2) = \left \{ \exp (\theta G)| \begin{pmatrix}
0 & -1\\ 
  1 & 0  
\end{pmatrix}, \theta \in \mathbb{R}/ 2 \pi \mathbb{Z} \right \},$ where $G$ is the infinitesimal operator of the group. The infinitesimal operator $G$ is thus encapsulating the group information. 

The group action, defined as $g_{\epsilon} \cdot x = \exp(\epsilon G) x$, that is, $\exp(\epsilon G)$ corresponds to the mapping induced by the action of the group element $g_{\epsilon}$ onto the data $x$.

One can exploit the Taylor series expansion of the exponential map to obtain its linearized version
\begin{align}
\label{eq:approx_lie_first}
    g \cdot x = \exp( \epsilon G) x  \approx (I+\epsilon G)x 
\end{align}  
where $x$ is the data and $\exp( \epsilon G) x $ its transformed version with respect to the group induced by the generator $G$. For more details regarding Lie group and the exponential map refer to \cite{hall2015lie}.

\section{The InfoNCE Framework}
\label{app:infoNCE}

We begin our analysis by considering the InfoNCE loss function, this loss function is commonly used in contrastive learning frameworks and has the benefit of providing multiple equivalent formulations that ease the derivation of insights regarding self-supervised learning embedding. In this section, we propose first 
a novel formulation of the InfoNCE which allows us to formalize common intuitions regarding contrastive learning losses. We then leverage this reformulation to provide an upper bound of the InfoNCE loss that will be central to our analysis regarding the role and properties of the projector. W.l.o.g. the proof will be first derived with $|| f_i^{(k)} ||=1$ to not introduce further notations.

The following proposition is a re-expression of the InfoNCE that allows us to consider this loss function as a regularized non-contrastive loss function where the regularizations provides insights into how this loss is actually behaving.

\begin{proposition}
\label{prop:infoNCE_entropy}
The infoNCE can be reformulated as
\begin{equation}
\label{eq:infoNCE_three}
\small{\mathcal{L}_{\mathrm{infoNCE}} = \frac{1}{N} \sum_{i=1}^{N} \left (-\beta  f_i^{(1)^T} f_i^{(2)} - \beta  f_i^{(1)^T} \mathbb{E}_{\mathcal{D}_{-i}}[\tilde{f}] +  H( \mathcal{D}_{-i} ) \right )}
\end{equation}
where, $H$ denotes the entropy, $\mathcal{D}_{-i}$ is a probability distribution such that $\textit{p}_{\mathcal{D}_{-i}}(\tilde{f}_l;\beta)= \frac{\exp(\beta f_i^{(1)^T} \tilde{f}_l )}{\sum_{k \in B_{-i}} \exp(\beta f_i^{(1)^T} \tilde{f}_k )}$ and recall that $\left \{\tilde{f}_l: l \in B_{-i} \right \} = \left \{ f_j^{(k)}: j\neq i, k=1,2 \right \}$ (Proof in App.\ref{proof:infoNCE_entropy}).
\end{proposition}
When minimizing Eq.~\ref{eq:infoNCE_three}, the first term forces the two augmented data in the projector space to be as similar as possible, which is the intuition behind most self-supervised learning losses. Note that only relying on such a similarity loss often leads to the collapse of the representation, that is one issue that is tackled in non-contrastive self-supervised learning using various hacks \cite{DBLP:journals/corr/abs-2006-09882,DBLP:journals/corr/abs-2006-07733}.
\paragraph{How InfoNCE selects its negative pairs?}
From Proposition~\ref{prop:infoNCE_entropy}, we observe that, when training under the InfoNCE, two additional terms act as a regularization to this non-contrastive similarity loss function, hindering the representation's collapse. In particular, the second term, $\beta  f_i^{(1)^T} \mathbb{E}_{\mathcal{D}_{-i}}[\tilde{f}]$, pushes the first augmented data in the projector space to be as different as possible to a weighted average of the other data. This expectation is dependant on the probability distribution $\mathcal{D}_{-i}$, which assigns high probability to data that are similar to $f_i^{(1)}$. Thus, the more the $f_i^{(1)}$ resembles $\tilde{f}_l$, the more they account for the loss, and hence the more they tend to be repelled. Now, it is clear that if the distribution $\mathcal{D}_{-i}$ is uniform, then all the data will tend to account for the same error. The third term avoids this by forcing this distribution to be with local support, and therefore, only consider a few instances of $\tilde{f}_l$, in particular, the ones that are the most similar to $f_{i}^{(1)}$. This last term thus enforces the $\tilde{f}_l$ to be different from each other by being repelled with different strength from $f^{(1)}_i$.

\paragraph{How does the temperature parameter affects these regularizations?}
From the Proposition~\ref{prop:infoNCE_entropy}, we develop some understanding regarding the temperature parameter $\beta$; in particular, this parameter influences the distribution $\mathcal{D}_{-i}$, which in turn, characterizes how local or global is the data taken into account into the repulsion process allowing to the differentiation between every data embedding. We first observe that, when $\beta \rightarrow +\infty$, $\mathbb{E}_{\mathcal{D}_{-i}}[\tilde{f}] = \argmax_{\tilde{f}_l, l \in B_{-i}} f_i^{(1)^T} \tilde{f}_{l}$, which induces that only the closest data from $f_i^{(1)}$ that is not $f_i^{(2)}$ is taken into account in the error. Besides, when $\beta \rightarrow \infty$, $H(\mathcal{D}_{-i}) \rightarrow 0$, therefore, this regularization term does not affect the loss function. This shows that increasing the temperature parameter reduces the support of the candidates for negative pairs and that the ones considered are the ones most similar to the considered data. Therefore, the temperature parameter effects the support of the per data distribution underlying the regularisation terms, $f_i^{(1)^T} \mathbb{E}_{\mathcal{D}_{-i}}[\tilde{f}]$ and $H( \mathcal{D}_{-i} ) $. From this, it is clear that having only one temperature parameter for the entire dataset is not adequate, as the locality of the distribution should be aligned with the volume of the data required to be repelled.

\section{Proofs}

\subsection{Proposition \ref{prop:infoNCE_entropy}}
\begin{proof}
\label{proof:infoNCE_entropy}
\begin{align}
\mathcal{L}_{\mathrm{infoNCE}} & = - \frac{1}{N} \sum_{i=1}^{N}  \log( \frac{\exp(\beta f_i^{(1)^T} f_i^{(2)}) }{\sum_{l \in B_{-i}} \exp(\beta f_i^{(1)^T} \tilde{f}_l) }).
\end{align}
Let's denote by $Z_i = \sum_{l \in B_{-i}} \exp(\beta f_i^{(1)^T} \tilde{f}_l)$ and let's consider the loss function for each data $i$ denoted by
$l_i = - \log ( \frac{ \exp(\beta f_i^{(1)^T} f_i^{(2)}) }{ Z_i})$.

Let's now define $\mathcal{D}_{-i}$ a probability distribution such that $\textit{p}_{\mathcal{D}_{-i}}(\tilde{f}_l;\beta)= \frac{\exp(\beta f_i^{(1)^T} \tilde{f}_l )}{Z_i}$.

We then compute the entropy of such distribution,

\begin{align}
    H(\mathcal{D}_{-i}) & = \sum_{l \in B_{-i}} \textit{p}_{\mathcal{D}_{-i}}(\tilde{f}_l;\beta) \log ( \frac{1}{\textit{p}_{\mathcal{D}_{-i}}(\tilde{f}_l;\beta)}) \nonumber \\
    & = \log(Z_i) - \beta \sum_{l \in B_{-i}} \textit{p}_{\mathcal{D}_{-i}}(\tilde{f}_l;\beta)  f^{(1)^T}_i \tilde{f}_l \nonumber \\
    & = \log(Z_i) - \beta f_i^{(1)^T} \mathbb{E}_{\mathcal{D}_{-i}} [ \tilde{f} ].
\end{align}
Thus, $\log(Z_i) = H(\mathcal{D}_{-i}) + \beta f_i^{(1)^T} \mathbb{E}_{\mathcal{D}_{-i}} [ \tilde{f} ].$

From $l_i = \log(Z_i) - \beta f_i^{(1)^T} f^{(2)}_i$ we obtain
\begin{equation*}
    l_i  = -\beta f_i^{(1)^T} ( f_i^{(2)} - \mathbb{E}_{\mathcal{D}_{-i}} [ \tilde{f} ]) + H(\mathcal{D}_{-i}).
\end{equation*}
Averaging over the $i=1,\dots,N$ concludes the proof.
\end{proof}

\subsection{Proposition \ref{prop:upperbound_linear}}

\begin{proof}
\label{proof:upperbound_linear}
We consider the result of Proposition~\ref{prop:infoNCE_entropy} and we take the derivative of $Z_i$ (described in Proof~\ref{proof:infoNCE_entropy}) with respect to $\beta$,

\begin{align}
    \frac{\delta}{\delta \beta} \log(Z_i) & = \frac{ \sum_{l \in B_{-i} } f^{(1)^T} \tilde{f}_l \exp( \beta f^{(1)^T} \tilde{f}_l) }{ \sum_{l \in B_{-i}} \exp( \beta f^{(1)^T} \tilde{f}_l)}  \nonumber \\
    & = f^{(1)^T} \frac{ \sum_{l \in B_{-i}}  \tilde{f}_l \exp( \beta f^{(1)^T} \tilde{f}_l) }{ \sum_{l \in B_{-i}} \exp( \beta f^{(1)^T} \tilde{f}_l )} \nonumber \\
    & = f^{(1)^T} \mathbb{E}_{\mathcal{D}_{-i}} [ \tilde{f} ].
\end{align}
Then,
\begin{equation*}
    \log(Z_i|_{\beta'=\beta}) - \log(Z_i|_{\beta'=0}) = f^{(1)^T}_i \int_{0}^{\beta} \mathbb{E}_{\mathcal{D}_{-i}}[\tilde{f}] d\beta'.
\end{equation*}
Now, given that $Z_i|_{\beta'=0} = \sum_{l \in B_{-i}} 1 = 2(N-1)$, we have

\begin{equation*}
    \log(Z_i|_{\beta'=\beta}) = f_i^{(1)^T} \int_{0}^{\beta} \mathbb{E}_{\mathcal{D}_{-i}}[\tilde{f}] d\beta' + \log(2(N-1)).
\end{equation*}
Therefore,
\begin{equation*}
    l_i = -f^{(1)^T}_i \int_{0}^{\beta} (f^{(2)}_i - \mathbb{E}_{\mathcal{D}_{-i}}[\tilde{f}]) d\beta' + \log(2(N-1))
\end{equation*}
Now, defining $\mathbb{E}_{\beta} [\mathbb{E}_{\mathcal{D}_{-i}}[\tilde{f}]] = \frac{1}{\beta} \int_{0}^{\beta} \mathbb{E}_{\mathcal{D}_{-i}(\beta')}[\tilde{f}] d\beta'$, we have

\begin{align}
l_i  & =  -\beta  f_i^{(1)^T} (f_i^{(2)} -  \mathbb{E}_{\beta} [\mathbb{E}_{\mathcal{D}_{-i}}[\tilde{f}]]  +  \log(2(N-1)) \nonumber \\ 
& =  -\beta  f_i^{(1)^T} f_i^{(2)} + \beta   \mathbb{E}_{\beta} [ f_i^{(1)^T} \mathbb{E}_{\mathcal{D}_{-i}}[\tilde{f}]]  +  \log(2(N-1)) \nonumber \\
& \leq -\beta  f_i^{(1)^T} f_i^{(2)} + \beta \mathbb{E}_{\beta} [ f_i^{(1)^T} \tilde{f}_i^{\star} ] + \log(2(N-1)) \nonumber \\
& = -\beta  f_i^{(1)^T} f_i^{(2)} + \beta \frac{1}{\beta} \int_{0}^{\beta} f^{(1)^T} \tilde{f}_i^{\star} d\beta' + \log(2(N-1)) \nonumber \\
& = -\beta  f_i^{(1)^T} f_i^{(2)} + \beta f^{(1)^T} \tilde{f}_i^{\star} + \log(2(N-1)) \nonumber \\
& = -\beta ( f_i^{(1)^T} ( f_i^{(2)} - \tilde{f}_i^{\star})) + \log(2(N-1)).
\end{align}

Now, recall that recall that for sake of clarity we used $||f_i^{(k)}||_2=1$.  Adding the normalization and assuming a linear projector, i.e., $g(h_i^{(k)}) = W^T h_i^{(k)}$ with $ W \in \mathbb{R}^{ d_{enc} \times d_{proj}}$  we obtain

\begin{align}
l_i = - \beta \frac{ h_i^{(1)^T} W W^T h_i^{(2)}}{|| W^T h_i^{(1)^T} || || W^T h_i^{(2)} ||} + \beta \frac{ h_i^{(1)^T} W W^T \tilde{h}_i^{\star}}{|| W^T h_i^{(1)^T} || || W^T \tilde{h}_i^{\star} ||} +\log(2(N-1))
\end{align}
Averaging over the $i=1,\dots,N$ concludes the proof.
\end{proof}

\subsection{Proposition \ref{prop:kernel_linear}}

\label{proof:kernel_linear}
\begin{proof}
\begin{align*}
    \mathcal{L}_{\text{invariance}} = \frac{1}{N} \sum_{i=1}^{N}  - \mathrm{Sim}(W^T \boldsymbol{h_i^{(1)}}, W^T (\boldsymbol{h_i^{(1)}} + \boldsymbol{v_i}  ) )
\end{align*}

is minimized if and only if $\forall i=1,\dots, N, \;\; \cos(\theta_{\boldsymbol{h_i^{(1)}}, \boldsymbol{h_i^{(1)}}+ \boldsymbol{v_i}})=1$, where $\theta_{\boldsymbol{h_i^{(1)}}, \boldsymbol{h_i^{(1)}}+\boldsymbol{v_i}}$ defines the angle between $W^T \boldsymbol{h_i^{(1)}}$ and  $W^T (\boldsymbol{h_i^{(2)}}+\boldsymbol{v_i})$. Denoting by $\propto$ the signed colinearity operator, we have
\begin{align}
    & \cos(\theta_{\boldsymbol{h_i^{(1)}}, \boldsymbol{h_i^{(1)}}+ \boldsymbol{v_i}})=1 \\
    & \iff  \theta_{\boldsymbol{h_i^{(1)}}, \boldsymbol{h_i^{(1)}}+ \boldsymbol{v_i}} = 0 \\
    & \iff W^T \boldsymbol{h_i^{(1)}} \propto W^T ( \boldsymbol{h_i^{(1)}} + \boldsymbol{v_i}) \\
    & \iff W^T \boldsymbol{v_i} = 0 \;\; \text{or} \;\; W^T \boldsymbol{h_i}^{(1)} \propto W^T \boldsymbol{v_i} \\
    & \iff W^T \boldsymbol{v_i} = 0 \;\; \text{or}  \;\; W^T \boldsymbol{h_i^{(1)}} \propto  W^T (\boldsymbol{h_{i}^{(2)}} - \boldsymbol{h_{i}^{(1)}}) 
\end{align}

Since, $\boldsymbol{h_i^{(1)}}$ is not colinear with $\boldsymbol{h_{i}^{(2)}}$, we obtain
\begin{align}
    \hspace{-3.6cm} \iff W^T \boldsymbol{v_i} = 0.
\end{align}
Therefore, $\boldsymbol{v}_i \in \mathrm{Ker}(W)$.
\end{proof}

\subsection{Proposition \ref{prop:rank_nullity}}
\label{app:rank_nullity}
\begin{proof}
We have that $\mathrm{Dim}(W) = \mathrm{Nullity}(W) + \mathrm{Rank}(W)$. Leveraging Proposition~.\ref{prop:kernel_linear}, we know that a projector minimizing $\mathcal{L}_{\text{invariance}}$ has the following property,
$\mathrm{Nullity}(W) \geq \mathrm{Dim}(\mathrm{Rank}(\boldsymbol{v_1},\dots,\boldsymbol{v_N})) $. As $(\boldsymbol{v_1},\dots,\boldsymbol{v_N}))$ is defined as the augmentation direction of the input manifold in the backbone encoder, it is clear that their rank is defined by the dimension of the subspace the encoder project the augmented directions. 
\end{proof}

\subsection{Proposition \ref{prop:kernel_nonlinear}}
\label{app:kernel_nonlinear}
\begin{proof}
A commonly used tool in the analysis of manifold induced by Lie group is the linearization of the exponential map. This exhibits the Lie algebra of the group, that is, a vector space to which the generator belongs (some details for in Appendix~\ref{app:liegroup}). We follow this approach and consider the linearized exponential map leading to
\begin{align}
     \mathcal{L}_{\text{invariance}} & = -\frac{1}{N} \sum_{i=1}^{N} \mathrm{Sim}(W^T \boldsymbol{h_i^{(1)}}, W^T (\boldsymbol{h_i^{(1)}} + \epsilon_i G \boldsymbol{h_i^{(1)}}  ) ).
\end{align}
Following the proof in Appendix~\ref{proof:kernel_linear}, we obtain that $\mathcal{L}_{\text{invariance}} $ is minimized if and only if $\mathrm{Col}(G) \in \mathrm{Ker}(W)$, where $\mathrm{Col}(G)$ denotes the columns of the generator $G$. Note that, if we lift the linearization and keep higher order term in the exponential map, we obtain that the powers of $G$ should be in the kernel of $W$.  
\end{proof}

\begin{figure}[t]
    \centering
    \includegraphics[width=.5\linewidth]{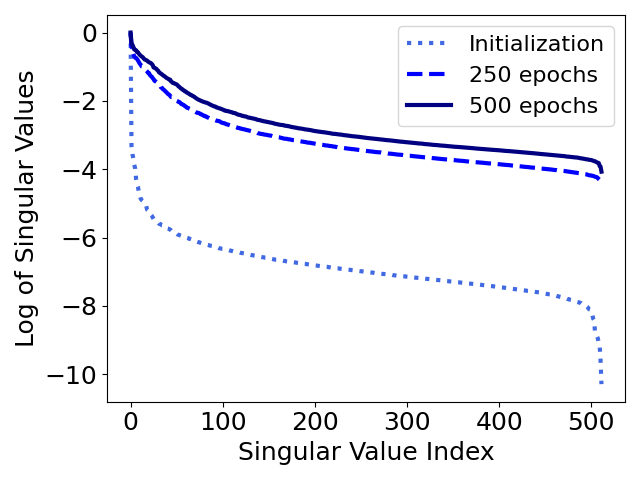}
    \caption{Encoder space log-singular values for the transformed Cifar$100$ dataset where the encoder space dimension $=512$ and a linear projector of output dimension $=128$. The log-spectrum is evaluated at different training time: initialization (\textit{dotted line}), half-training time (\textit{dashed line}), and after training (\textit{solid line}). In these three settings, the calculation of the rank following Eq.~\ref{eq:rank}, leads to $\mathrm{Rank}_{\tau}\left [ h^{(1)}(x_1), \dots, h^{(1)}(x_N) \right ] = 512$, that is, the representation induced by the encoder is full rank (and close to full rank at the initialization) until the end of the training as for the case where the input data are not transformed, in Fig.~\ref{fig:rank_enc_cifar_linear}. 
    }
    \label{fig:rank_enc_transformed_cifar_linear}
\end{figure}

\section{Toy-example -  Rank versus Augmentation Strength}
\label{app:rank_vs_rot_toy}
\begin{figure}[ht]
    \centering
    \includegraphics[width=.5\linewidth]{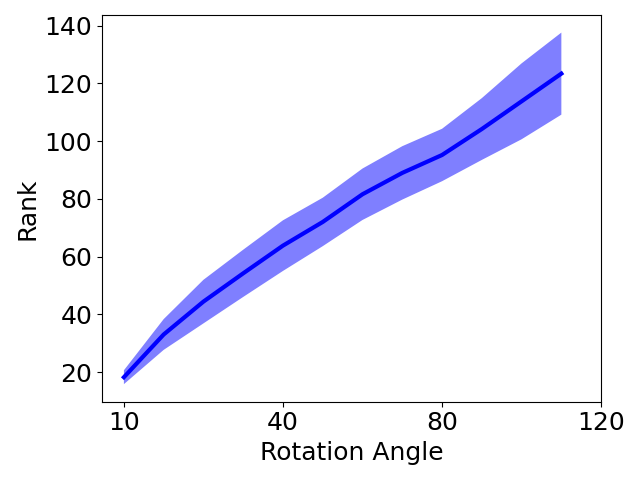}
     \vspace{-.5cm}
    \caption{\textbf{Rank of the covariance matrix of $500$ rotated version of a random one-hot image} of dimension $32 \times 32$ (mean and standard deviation for $5$ runs). We sample the angle between $0$ and the $x$-axis angle value uniformly. The larger the transformation, the higher rank the collection of images is, and there the lower rank a linear map is required to provide an invariant map to this specific transformation}
    \label{fig:rank_vs_rot_toy}
\end{figure}

\end{document}